# Directed Domain Fine-Tuning: Tailoring Separate Modalities for Specific Training Tasks

DANIEL WEN*, University of California, Santa Cruz, USA
NAFISA HUSSAIN, University of California, Santa Cruz, USA

Large language models (LLMs) and large visual language models (LVLMs) have been at the forefront of the artificial intelligence field, particularly for tasks like text generation, video captioning, and question-answering. Typically, it is more applicable to train these models on broader knowledge bases or datasets to increase generalizability, learn relationships between topics, and recognize patterns. Instead, we propose to provide instructional datasets specific to the task of each modality within a distinct domain and then fine-tune the parameters of the model using LORA. With our approach, we can eliminate all noise irrelevant to the given task while also ensuring that the model generates with enhanced precision. For this work, we use Video-LLaVA to generate recipes given cooking videos without transcripts. Video-LLaVA's multimodal architecture allows us to provide cooking images to its image encoder, cooking videos to its video encoder, and general cooking questions to its text encoder. Thus, we aim to remove all noise unrelated to cooking while improving our model's capabilities to generate specific ingredient lists and detailed instructions. As a result, our approach to fine-tuning Video-LLaVA leads to gains over the baseline Video-LLaVA by 2% on the YouCook2 dataset. While this may seem like a marginal increase, our model trains on an image instruction dataset 2.5% the size of Video-LLaVA's and a video instruction dataset 23.76% of Video-LLaVA's.

Additional Key Words and Phrases: Large Vision Language Models, Multimodality, Fine-tuning, Large Language Models, Visual Question Answering, Instruction Fine-Tuning

## 1 INTRODUCTION

Most recently, the advent of multimodal models has significantly expanded the frontiers of Artificial Intelligence research and application, showcasing remarkable versatility and efficiency across many tasks that were once the exclusive domain of single-modality large language models (LLMs). Unlike their predecessors, which primarily processed and generated text-based information, multimodal models are adept at understanding, interpreting, and generating outputs that span across multiple forms of data, including but not limited to text, images, audio, and video. This capability not only enhances the models' comprehension and interaction with the world but also aligns more closely with human cognitive processes, which are inherently multimodal. Integrating multiple data types allows these models to perform a broader spectrum of tasks with enhanced accuracy, creativity, and contextual understanding, thus opening new avenues for AI applications in more nuanced and complex scenarios.

Multimodal models have found application in diverse tasks, illustrating their flexibility and the breadth of their utility. Some of the most notable applications include visual question answering (VQA), where the model responds to text-based questions about visual content; image captioning, which involves generating descriptive text for images; and audio-visual speech recognition, where the model uses both visual and auditory signals to improve speech recognition accuracy. Additionally, these models have been successfully employed in sentiment analysis by analyzing text and vocal tones, in object detection and recognition by integrating spatial data with visual inputs, and in machine translation, where visual context enhances the accuracy of text translations.

Another application of multimodal models is the generation of step-by-step instructions for complex tasks. This involves not just understanding a text-based query but also analyzing accompanying visual data to produce a coherent, easily understandable sequence of instructions. A multimodal model needs to have a sufficient temporal understanding of events and actions taking place within video inputs. Models should also be able to capture a detailed understanding of the relationships among different actions presented. Additionally, the reasoning ability to work through the relationships between different tools, ingredients, or materials is necessary for a meaningful understanding of inputted tasks. This capability represents a significant leap forward in models' potential to assist with practical, everyday tasks, offering personalized guidance that considers the unique aspects of each scenario.

---

*Both authors contributed equally to this research.

Authors' addresses: Daniel Wen, dywen@ucsc.edu, University of California, Santa Cruz, CA, USA; Nafisa Hussain, University of California, Santa Cruz, USA, nahussai@ucsc.edu.

We propose a new approach to fine-tune Large Vision Language Models to the task of step-by-step instruction generation. Through the selected domain of Recipe Generation, we fine-tune Video-LLaVA-7B to generate thorough step-by-step recipes and a list of ingredients with specific measurements for cooking videos that contain no transcripts or auditory information of the contents of the video [2]. We specifically fine-tune for each modality Video-LLaVA accepts as input on a different task related to recipe generation and cooking activities. Our experiments show optimistic results in the work of fine-tuning modalities models on distinct tasks for developing a comprehensive understanding of detailed multi-step procedures.

## 2 RELATED WORK

### 2.1 Large Language Models

In recent years, the artificial intelligence community has made massive advancements in various domains through its ability to emulate human intelligence. Among the most groundbreaking developments in AI are large language models (LLMs), such as Gemini AI, and its open-source alternative, ChatGPT [1] [3]. More specifically, LLaMA and Vicuna utilize multimodal learning to enhance LLM understanding and generation abilities [4] [5]. Using the language-based multimodal pretraining framework, LanguageBind, video-language pretraining can be extended to multiple modalities for the LLMs in Video-LLaVA's architecture, which is the framework that our model further builds upon [8] [6].

### 2.2 Large Vision Language Models

While the previously mentioned LLMs pertain to text, multi-modality opens new opportunities for other forms of inputs, such as video. Large vision language models (LVLMs) merge fields of computer vision and natural language processing to comprehend visual and textual information, which is the field our model is aimed at. Video-ChatGPT is an LVLM extension of the GPT architecture, leveraging LLaVA as its foundation [6]. It employs CLIP ViT-L/14 as its visual encoder and Vicuna as its large language model, the same encoder and decoders used in Video-LLaVA [11]. However, when testing Video-ChatGPT's performance in recipe generation, it does not accurately understand the steps and ingredients in the provided cooking videos. Similarly, Video-LLaVA outperformed VideoChat2, another multimodal LVLM tailored for video chat applications, in the specific domain of recipe generation. Hence, our objective is to meticulously fine-tune each input of Video-LLaVA's multimodal architecture to a distinct domain- recipe generation in our example [7]. We aim to enhance the model's comprehension of cooking videos significantly, enabling it to generate precise ingredient lists and detailed procedural instructions.

### 2.3 Zero-Shot Anticipation

In search of cooking-related images and videos to fine-tune Video-LLaVA, we came across the Tasty Video dataset, a diverse set of 2511 different cooking videos with recipes for zero-shot learning, recognition, and anticipation. Sener and Yao utilize this dataset to test their model's anticipation abilities [13]. Their model architecture comprises a sentence encoder, recipe RNN, and sentence decoder jointly trained end-to-end. The sentence encoder is a bi-directional LSTM followed by a max pooling over each dimension of the hidden units, and the sentence decoder utilizes an LSTM-based neural language model to transform the representations back into coherent sentences. The recipe RNN is an LSTM that takes in the recipe steps as fixed-length representations as inputs, and each hidden state vector is conditioned on the previous steps; this enables the recipe RNN to predict recipe steps using its previous steps. The model showcases similar zero-shot anticipation predictions for recipe instructions that we aim to achieve as well. While their approach learns instructional tasks from text and then transfers the knowledge to the video, ours is to fine-tune text, image, and video inputs of an LVLM and attain precise instructions.

## 3 APPROACH

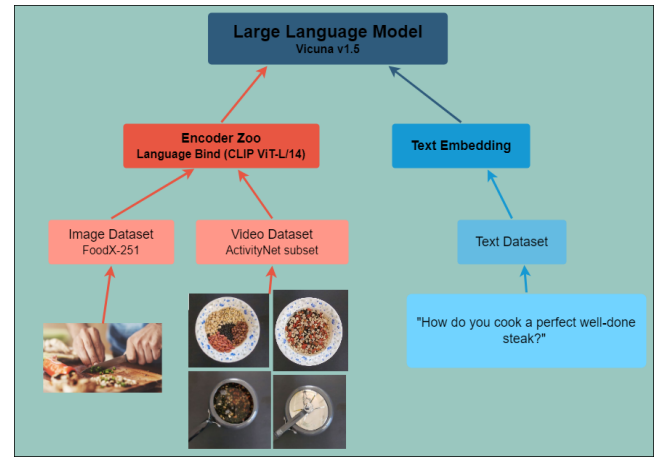

Fig. 1

Our approach is to use LORA parameter efficient fine-tuning (PEFT) on Video-LLaVA's multimodal architecture; we then provide instructional datasets specific to each modality to a specific task within the distinct domain [9]. The image modality trains on the FoodX-251 dataset, with 251 fine-grained food categories and 158k food images [10]. Video-LLaVA trains on 2511 videos from the Tasty dataset containing over-the-head 3rd-person recipe tutorial videos. Finally, the text modality trains on 4000 general cooking questions we generated with **gpt-3.5-turbo**. We found that we generated better recipe instructions when training the text modality on general cooking questions rather than questions about specific dishes. As seen in Figure 1, we use LanguageBind to map the image and video modalities into the textual feature space, so the model can learn from a unified visual representation; this is then fed into the LLM, Vicuna. Compared to baseline Video-LLaVA, which trains on a 665k image-text instruction dataset and a 100k video-text instruction dataset, our model's image instruction dataset is 23.76% in size, and our video instruction dataset is 2.5% in size.

Below is the JSON format for the image encoder. Notice how we only ask what the dish is.

```
{
  "id": "1234567890",
  "image": "file_name.jpg",
  "conversations": [
    {
      "from": "human",
      "value": "What is the dish in this image?"
    },
    {
      "from": "gpt",
      "value": "<Dish name>"
    }
  ]
}
```

Below is the JSON format for the video encoder. Not only do we ask for the recipe from the dish in the provided video, but we also ask for specific measurements to ensure that the model outputs quantitative values. Previous tests on other models, such as Video-ChatGPT or VideoChat2, show that this prompt tuning is necessary; otherwise, the model is more likely to output values like "a small amount of flour" instead of "1 cup of flour".

```
{
  "id": "0",
  "video": "video_name.mp4",
  "conversations": [
    {
      "from": "human",
      "value": "Can you give me a recipe from the
          provided videos and include specific
          measurements for each of the
          ingredients?"
    },
    {
      "from": "gpt",
      "value": "<The recipe of the given video.>"
    }
  ]
}
```

Below is the JSON format for the text encoder. This is an example of a random general cooking question, which we generated 4000 of in order to tune the NLP capabilities of our model. The generated answer is cut off for the example.

```
{
  "id": "0",
  "model": "",
  "conversations": [
    {
      "from": "human",
      "value": "What is the best way to cook a
          juicy steak?"
    },
    {
      "from": "gpt",
      "value": "The best way to cook a juicy steak
          is to start by seasoning the steak
          with salt and pepper and allowing it to
          come to room temperature. Preheat a
          cast iron skillet over high heat and
          ..."
    }
  ]
}
```

## 4 RESULTS

### 4.1 Evaluation Setup

We construct a step-by-step instruction generation dataset based on cooking videos and recipes from the YouCook2 dataset [12]. By default, the dataset consists of 2000 long untrimmed videos from 89 cooking recipes; on average, each distinct recipe has 22 videos. The procedure steps for each video are annotated with temporal boundaries and described by imperative English sentences. The videos were downloaded from YouTube and are all in the third-person viewpoint. All the videos are unconstrained and can be performed by individuals at their houses with unfixed cameras. We parsed through the associated recipes from each video to create a dataset of questions asking for a step-by-step procedure for each recipe and specific measurements for all the ingredients to be included. Our final dataset contained 408 videos from 62 different cooking recipes. Compared to baseline Video-LLaVA, which trains on a 665k image-text instruction dataset and a 100k video-text instruction dataset, our model's image instruction dataset is 23.76% in size, and our video instruction dataset is 2.5% in size.

To enable automatic evaluation, we used **gpt-3.5-turbo** to score responses from Video-LLaVA on the constructed dataset on a scale from one to five with a score in the 3.5-5 range counted as 'Yes' accurate responses to the ground truth and score of 1-3.4 counted as 'No' inaccurate responses to the ground truth. The evaluation scores are based on finding a meaningful match between the generated responses and ground truth answers. This includes assessing the similarity between ingredient lists and measurements, the order of actions within a recipe, and the similarity of using ingredients within procedure steps. Accuracy is computed as the count of 'Yes' responses generated by **gpt-3.5-turbo** over the total number of responses generated over the evaluation dataset.

Below is the system prompt fed to **gpt-3.5-turbo** to evaluate responses generated by fine-tuned Video-LLaVA and baseline Video-LLaVA on the YouCook2 dataset subset based on the specific criteria we outlined above.

```
messages=[{
            "role": "system",
            "content":
                "You are an intelligent chatbot
                    designed for evaluating the
                    correctness of generative
                    outputs for question-answer
                    pairs. "
                "Your task is to compare the
                    predicted answer with the
                    correct answer and determine
                    if they match meaningfully.
                    Here's how you can accomplish
                     the task:"
                "------"
                "##INSTRUCTIONS: "
                "- Focus on the meaningful match
                    between the predicted answer
                    and the correct answer.\n"
                "- Consider synonyms or
                    paraphrases as valid matches
                    .\n"
                "- Evaluate the correctness of the
                     prediction compared to the
                    answer."
                "- Consider the similarity between
                     ingredient lists and
                    measurements."
        }]
```

### 4.2 Results Analysis

To expand further on our findings from fine-tuning Video-LLaVA's multiple modalities on distinct tasks from the target task, we delve deeper into the specific outcomes observed through our experiments. Our investigation centered on comparing the quality of generated step-by-step recipes of baseline Video-LLaVA-7B and fine-tuned Video-LLaVA-7B on each modality for the domain of recipe generation. Our experimental results demonstrate that our method of fine-tuning Video-LLaVA leads to significant boosts in generating precise recipe instructions over general-purpose trained Video-LLaVA. This improvement is quantitatively evident when examining the accuracy comparisons of fine-tuned Video-LLaVA and baseline Video-LLaVA on the constructed YouCook2 datasets in Table 1. The model subjected to fine-tuning performed with higher accuracy and received an overall higher average score over the baseline Video-LLaVA model.

One observation we made that could explain this finding was that baseline Video-LLaVA would tend to list off significantly more ingredients than actually needed in the ingredient listing portion of the recipe. fine-tuned Video-LLaVA tended to produce more concise recipe steps than baseline Video-LLaVA, whose responses were often more lengthy.

There is room for more robust evaluation through employing human evaluation on a subset of responses from both fine-tuned Video-LLaVA and baseline Video-LLaVA. We can gather a group of people to blindly judge between the recipes generated by both models which response is more accurate to a provided ground truth for a subset of responses produced by both models. Additionally, we can also break down the evaluation into 2 stages: One, where the ingredient listings with their specific measurements are first compared, and two, where the recipe instructions are solely compared.

Table 1. Comparison of Fine-Tuned Video-LLaVA and Baseline Video-LLaVA

| Metric | Fine-Tuned Video-LLaVA | Baseline Video-LLaVA |
| --- | --- | --- |
| Yes Count | 191 | 183 |
| No Count | 217 | 225 |
| Accuracy | 46.813% | 44.852% |
| Average Score | 3.1296 | 2.9223 |

### 4.3 Ablation Study: Temporal Understanding

We further investigated the validity of responses from Video-LLaVA in the temporal understanding of video content. Specifically, we explored Video-LLaVA's ability to consider prompts about time stamps. We were curious about the model's ability to ground generated steps factual timestamps from the input video to validate the proposed instructions on specific video sections.

Invoking the model to answer in response to timestamps requires the model to interpret timestep tokens for further specific embeddings. Timestamps can be considered an additional modality of understanding for models. Understanding time measurements in relation to the content of videos requires a separate joint embedding of video tokens and timestamp elements in conjunction. Currently, Video-LLaVA does not support timestamp embeddings for video inputs. Hence, the model's reasoning process through the timestamp question is undeveloped. When asked about timestamps where certain steps from the generated recipes show up in the input video, Video-LLaVA tends to respond with clearly inaccurate timestamp estimates. Timestamps generated for steps that are queried about often tend to exceed the actual length of the video. Multiple queries about where in the video separate distinct steps in the procedure occur result in conflicting responses. Responses will include overlapping timestamps or identical timestamps citing steps that occur in a clearly sequential ordering occur in parallel.

## 5 CONCLUSION AND FUTURE WORK

### 5.1 Conclusion

In conclusion, our research demonstrates a significant advancement in utilizing Large Language Models (LLMs) and Large Visual Language Models (LVLMs) within the specific context of generating culinary recipes from cooking videos. By adopting a tailored approach that involves training these models with modality-specific instructional datasets within the culinary domain, followed by fine-tuning with LORA, we have markedly enhanced model precision and effectiveness. This methodology effectively filters out irrelevant noise, thereby ensuring that the generated content is not only more accurate but also highly relevant to the task at hand. The use of Video-LLaVA, with its multimodal architecture, has been instrumental in achieving these results, allowing for seamless integration of cooking images, videos, and text inputs. Our work not only showcases the potential of fine-tuned LLMs and LVLMs in specialized domains but also sets a precedent for future research to leverage artificial intelligence for task-specific knowledge extraction and content generation. This study underscores the importance of domain-specific customization in developing AI models and opens avenues for exploring similar methodologies across various fields.

### 5.2 Future Work

One avenue of further investigation is grounding generated procedural steps on keyframes from the input video. Tracing through Video-LLaVA's self-attention scoring placed on selected embedded video frames to select frames with the highest attention values would be useful in providing visual aid to generated steps. This can be conducted by adding hooks to capture attention weights during forward passes and averaging over-calculated scores. Further work can also be conducted in building out Video-LLaVA to accommodate time stamp embeddings for procedure generation grounded on time stamps from a video. Building out an encoder to process time stamp inputs and jointly align these inputs with video token embeddings will be useful in providing some basis for the model to extract meaning out of queries regarding time stamps and other forms of temporal understanding required. Additionally, these embeddings would enable us to apply a supervised approach to fine-tuning Video-LLaVA on videos and timestamp annotations to learn temporal patterns in procedures.

This current proposal focuses on procedure generation in the recipes domain. We hope to adapt this fine-tuning approach to more domains, such as step-by-step instruction generation for medical and manufacturing procedures. This would involve domain-specific instruction tuning and domain knowledge tuning into Video-LLaVA with high-quality dataset curation per specified domain. Dataset curation in this experiment for the domain of recipe generation was difficult due to the lack of Recipe datasets that include videos and corresponding recipes. Further manual construction is necessary to describe recipes for existing cooking videos to generate more datasets to evaluate this specific approach on the specific domain task. As mentioned earlier, our model's image instruction dataset is 23.76% in size of baseline Video-LLaVA's 665k image-text instruction dataset, and our video instruction dataset is 2.5% in size of baseline Video-LLaVA's 100k video-text instruction dataset.

From a development perspective, we hope to containerize our current workflow that runs on an 80 GB H100 GPU hosted on an AWS Lambda instance. This server instance has been quite useful throughout this project but has also incurred quite the cost. We wish to create a Docker image of our development environment to simplify the infrastructure setup process for different server architectures. We also hope to look into more fine-grained batch processing at the fine-tuning stage to speed up the process. Additionally, we hope to build a user interface with Gradio to accommodate the display of the generated ingredient listings and recipes, along with their corresponding video frame groundings.

## 6 MEMBER CONTRIBUTIONS

In this collaborative research effort, each author contributed distinct efforts to address the challenge of domain-specific instruction generation in Large Vision Large Language Models. Their contributions are outlined as follows.

Nafisa's contributions include:

- Conducting extensive literature review into potential training and evaluation datasets for step-by-step recipe generation tasks.
- Setting up workflow infrastructure in AWS Lambda instance with H100 GPU, installing correct dependencies, and setting correct system environment paths.
- Responsible for monitoring and adjusting fine-tuning parameters and batch training based on results over several inference cycle analyses
- Adapted Video-LLaVA's original evaluation criteria to suit the goals of our domain-specific task

Daniel's contributions include:

- Evaluated VideoChat2 zero-shot performance.
- Developed a script for image labeling of the FoodX-251 dataset.
- Developed a script for label-to-class name matching.
- Analyzed the performance of randomly generated questions and recipes.
- Generated a dataset of 4,000 cooking-related questions to fine-tune the NLP capabilities of Video-LLaVA.
- Built approach diagram
- Explored retrieval-augmented generation (RAG) as an approach to enhance the performance of Video-LLaVA's frame retrieval.
- Reviewed the attention mechanism employed by Video-LLaVA for potential use for frame retrieval.